\documentclass[10pt,twocolumn]{article}

\usepackage{times}
\usepackage{epsfig}
\usepackage{amssymb}
\usepackage{amsmath}
\usepackage{lineno}
\usepackage{authblk}
\usepackage{hyperref}
\usepackage{multirow}
\usepackage{booktabs}
\usepackage{xcolor}
\usepackage{colortbl}
\usepackage{adjustbox}
\usepackage{graphicx}
\usepackage{subcaption}
\usepackage{pifont}
\usepackage[T1]{fontenc}

\newcommand{\cmark}{\ding{51}}
\newcommand{\xmark}{\textcolor{black!35}{\ding{55}}}

\hypersetup{
  colorlinks=true,
  linkcolor=blue,
  citecolor=blue,
  urlcolor=blue,
  pdfborder={0 0 0},
}

\title{Distinguishing Visually Similar Actions: Prompt-Guided Semantic Prototype Modulation for Few-Shot Action Recognition}

\author[1]{Xiaoyang Li}
\author[1]{Mingming Lu\thanks{Corresponding author. Email: mingminglu@csu.edu.cn}}
\author[1]{Ruiqi Wang}
\author[1]{Hao Li}
\author[1]{Zewei Le}

\affil[1]{School of Computer Science and Engineering, Central South University, Changsha 410083, Hunan Province, China}

\date{}

\begin{document}
\maketitle

\begin{abstract}
Few-shot action recognition aims to enable models to quickly learn new action categories from limited labeled samples, addressing the challenge of data scarcity in real-world applications. Current research primarily addresses three core challenges: (1) temporal modeling, where models are prone to interference from irrelevant static background information and struggle to capture the essence of dynamic action features; (2) visual similarity, where categories with subtle visual differences are difficult to distinguish; and (3) the modality gap between visual-textual support prototypes and visual-only queries, which complicates alignment within a shared embedding space. To address these challenges, this paper proposes a CLIP-SPM framework, which includes three components: (1) the Hierarchical Synergistic Motion Refinement (HSMR) module, which aligns deep and shallow motion features to improve temporal modeling by reducing static background interference; (2) the Semantic Prototype Modulation (SPM) strategy, which generates query-relevant text prompts to bridge the modality gap and integrates them with visual features, enhancing the discriminability between similar actions; and (3) the Prototype-Anchor Dual Modulation (PADM) method, which refines support prototypes and aligns query features with a global semantic anchor, improving consistency across support and query samples. Comprehensive experiments across standard benchmarks, including Kinetics, SSv2-Full, SSv2-Small, UCF101, and HMDB51, demonstrate that our CLIP-SPM achieves competitive performance under 1-shot, 3-shot, and 5-shot settings. Extensive ablation studies and visual analyses further validate the effectiveness of each component and its contributions to addressing the core challenges. The source code and models are publicly available at \href{https://github.com/hnlxy/CLIP-SPM}{GitHub}.
\end{abstract}

\section{Introduction}\label{sec1}

Video action recognition~\cite{li2025manet,zhou2025efficient,li2024action,wang2025action}, a foundational topic in computer vision, has made significant progress. However, its effectiveness hinges on access to large and high-quality annotated datasets. This data-intensive paradigm faces fundamental constraints in practice: on the one hand, meticulous video annotation demands significant human and time resources; on the other hand, action categories in real-world scenarios follow a long-tailed distribution~\cite{perrett2023use,zhang2025causality,pang2024long,hu2025long}. Collecting sufficient samples for numerous emerging or rare categories, such as specific medical procedures or industrial anomalies, is therefore difficult, leading to a notable decline in model generalization. Consequently, exploring data-efficient learning paradigms has become pivotal for advancing action recognition toward practical deployment.

The field of few-shot action recognition (FSAR) has emerged to address this challenge by emulating human ability to rapidly acquire new concepts from limited examples. FSAR often employs meta-learning frameworks to extract cross-task and general knowledge, enabling accurate recognition of unseen actions with minimal training data. Despite early advances, contemporary FSAR methods still face three major challenges: (1) temporal modeling, where models are prone to interference from irrelevant static background information, making it difficult to capture the essential dynamic motion features; (2) visual similarity, where action categories with subtle visual differences (e.g., ``Running'' vs. ``Jogging'') are hard to distinguish, especially in few-shot settings; and (3) the modality gap, where query sets lack textual priors during the meta-testing phase, complicating the alignment between support and query representations, as visual features alone struggle to fully capture the semantic concepts.

Existing works have attempted to tackle these issues from multiple perspectives. For temporal modeling, OTAM~\cite{cao2020few}, which builds on dynamic time warping, aligns support and query sequences to address action speed variation, yet struggles to capture consistent motion information. TRX~\cite{perrett2021temporal} captures temporal relationships via self-attention but is prone to overfitting with limited samples. Regarding visual similarity, several approaches~\cite{wang2022hybrid,wang2024hyrsmpp} adopt feature augmentation strategies or devise complex distance metrics, yet they often overlook the potential of motion and semantic information. With the rise of vision-language models~\cite{radford2021learning}, some works~\cite{wang2025actionclip,rasheed2023fine,wang2024clip,cao2024exploring} have sought to leverage CLIP~\cite{radford2021learning}'s strong semantic priors. However, these methods generally treat text as fixed classification weights, rather than enabling deep interaction between visual content and linguistic semantics. Additionally, to mitigate the modality gap, CLIP-CPM$^2$C~\cite{guo2025consistency} attempts to compensate for missing query semantics by learning random text embeddings. However, the resulting embeddings often fail to effectively correlate with the visual content, hindering the ability to bridge the modality gap.

To address the above challenges, this paper proposes a cross-modal learning-based FSAR framework. Our main contributions are as follows:

\begin{itemize}
    \item We present CLIP-SPM, a cross-modal framework that jointly tackles three central FSAR challenges: robust temporal modeling, fine-grained visual similarity, and the support-query modality gap.
    \item We design a Hierarchical Synergistic Motion Refinement (HSMR) module that explicitly extracts both shallow and deep motion features, enforcing consistency between these levels to capture stable motion patterns under scarce supervision.
    \item We develop a Semantic Prototype Modulation (SPM) strategy, which learns query-relevant textual prompts from episode context to bridge the modality gap. SPM modulates the visual features by leveraging the semantic guidance from textual features, enhancing the discriminability between visually similar actions.
    \item We introduce a Prototype-Anchor Dual Modulation (PADM) method, which jointly refines support-side prototypes and aligns query features to a global anchor, yielding stronger cross-set consistency, and more discriminative prototypes.
\end{itemize}

We conduct extensive experiments on five public benchmarks, including Kinetics, SSv2-Full, SSv2-Small, UCF101, and HMDB51. The results demonstrate that CLIP-SPM achieves competitive performance under 1-shot, 3-shot, and 5-shot settings. Comprehensive ablation studies and visual analyses further validate the effectiveness and necessity of each component within CLIP-SPM.

The remainder of this paper is organized as follows: Section~\ref{sec2} reviews related work; Section~\ref{sec3} details the proposed CLIP-SPM; Section~\ref{sec4} presents experimental results and analysis; finally, Section~\ref{sec5} concludes the paper.

\section{Related work}\label{sec2}

This work is closely related to few-shot action recognition, temporal modeling, vision-language priors, and prototype refinement. We briefly review each line of work below.

\subsection{Few-shot action recognition}

Few-shot action recognition aims to recognize novel actions from only a few labeled examples. Classical metric-based methods, such as Prototypical Networks~\cite{snell2017prototypical}, construct class prototypes in an embedding space and classify queries by nearest-prototype matching. Video-based extensions, including CMN~\cite{zhu2018compound} and CMN-J~\cite{zhu2020label}, introduce memory mechanisms and semi-supervised strategies to better exploit scarce labeled and unlabeled clips, while ARN~\cite{zhang2020few} improves robustness to ordering and label noise. Although these approaches provide a solid foundation, they primarily emphasize appearance features and simple prototype matching, and thus struggle with complex temporal dynamics, modality discrepancies between visual and textual information, and the need for more discriminative prototypes under fine-grained action similarity. These limitations have motivated subsequent lines of work on temporal modeling, vision-language priors, and prototype refinement, which we review in the following subsections.

\subsection{Temporal modeling}

Temporal modeling is crucial in few-shot action recognition as it addresses speed and phase variations, ensuring the model can effectively recognize actions despite temporal differences. Early methods, such as OTAM~\cite{cao2020few}, align support and query sequences via dynamic time warping to improve temporal robustness. Later works, including TRX~\cite{perrett2021temporal}, HyRSM~\cite{wang2022hybrid}, and STRM~\cite{thatipelli2022spatio}, enhance this with relational and attention-based mechanisms to capture multi-scale temporal dynamics. Additionally, approaches like MTFAN~\cite{wu2022motion}, TA2N~\cite{li2022ta2n}, and HCR~\cite{li2024hierarchical} improve temporal alignment and aggregation, while MoLo~\cite{wang2023molo} focuses on motion-salient segments to reduce background interference. Despite these advances, most methods rely primarily on visual cues and implicitly model temporal dynamics, which can lead to insufficiently captured motion patterns. Our proposed HSMR module addresses this limitation by explicitly modeling both shallow and deep motion features and enforcing consistency between these levels. This approach improves the robustness of temporal representations, particularly in few-shot settings.

\subsection{Vision-language priors}

Vision-language models have become a powerful tool for injecting semantic priors into few-shot action recognition. By leveraging joint image-text representations, methods like CLIP~\cite{radford2021learning} significantly enhance the generalization ability of action recognition, particularly under few-shot settings where training data is scarce. CLIP-FSAR~\cite{wang2024clip} refines prototypes by incorporating textual descriptions, improving the discriminability of action categories. Similarly, STEN~\cite{chen2024cross} combines vision-language priors with spatio-temporal reasoning to boost FSAR performance. Methods like CLIP-M$^2$DF~\cite{guo2024multi} further explore multi-modal fusion, distilling knowledge from both vision and text across multiple views. CLIP-CPM$^2$C~\cite{guo2025consistency} enhances feature alignment by introducing consistency-driven learning with CLIP's vision-language priors, while CLIP-MEI~\cite{deng2025clip} integrates motion-enhanced inference to leverage temporal cues alongside textual semantics. Although these approaches benefit from CLIP's rich semantic priors, challenges remain in addressing the semantic ambiguity. Our proposed SPM strategy aims to bridge these gaps by dynamically learning query-relevant prompts from the episode context and integrating them with visual features, thus enhancing semantic discriminability under few-shot conditions.

\subsection{Prototype refinement}

Prototype refinement is crucial for few-shot action recognition, and recent methods have introduced various strategies to enhance class prototypes. Jiang et al.~\cite{jiang2024dual} propose a dual-prototype network that combines class and query prototypes, enabling more accurate class representation under few-shot conditions. Similarly, MoLo~\cite{wang2023molo} enhances prototypes by focusing on motion-salient fragments and applying contrastive learning over both long and short temporal segments, improving action recognition by reducing background interference. Additionally, HCR~\cite{li2024hierarchical} improve prototype consistency by integrating features at multiple levels, strengthening intra-class cohesion. TEAM~\cite{lee2025temporal} refines prototypes without explicit temporal alignment, instead leveraging unsupervised video segment matching to improve the robustness of prototypes to temporal misalignments. Despite these advancements, challenges persist in balancing high intra-class variance with background noise, and in combining multiple feature sources for prototype representation. Our proposed PADM method addresses these issues by refining both support-side prototypes and query features simultaneously, enhancing consistency across samples while preserving class discriminability.

\begin{figure*}[htb]
  \centering
  \includegraphics[width=\textwidth]{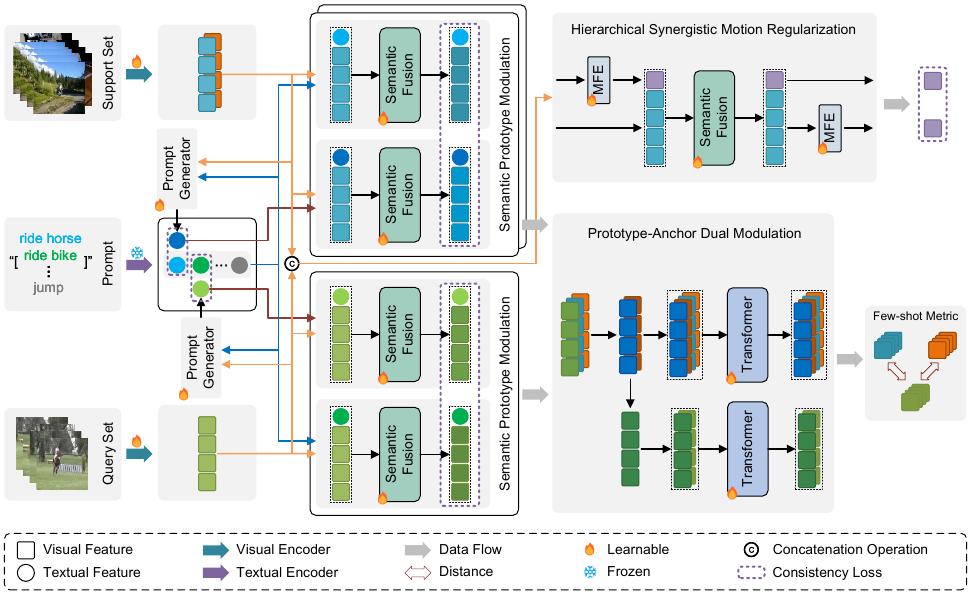}
  \caption{The architecture of CLIP-SPM. The Hierarchical Synergistic Motion Refinement (HSMR) extracts shallow/deep motion cues and enforces consistency, while the Semantic Prototype Modulation (SPM) provides semantics-guided feature modulation. The Prototype--Anchor Dual Modulation (PADM) jointly refines support prototypes and aligns query features with a global anchor through prototype and anchor modulation.}
  \label{fig1}
\end{figure*}

\section{Methodology}\label{sec3}

In this section, we present the proposed CLIP-SPM framework in detail. We begin by formalizing the few-shot action recognition problem, followed by an overview of the overall architecture. We then describe the three core components: Hierarchical Synergistic Motion Regularization (HSMR), Semantic Prototype Modulation (SPM), and Prototype--Anchor Dual Modulation (PADM). Finally, we introduce the loss functions used to jointly optimize the entire framework.

\subsection{Problem formulation}
\label{subsec3.1}

We adopt the standard few-shot learning paradigm with episodic training~\cite{snell2017prototypical} and cast FSAR as a meta-learning problem. The dataset is partitioned into disjoint subsets with non-overlapping label spaces: a training set $\mathcal{D}_{\text{train}}$ for meta-training and a test set $\mathcal{D}_{\text{test}}$ for meta-testing. The objective is to leverage knowledge acquired on $\mathcal{D}_{\text{train}}$ to rapidly generalize to unseen categories in $\mathcal{D}_{\text{test}}$.

Learning proceeds as a sequence of episodes (meta-tasks), each simulating a few-shot scenario. In an $N$-way $K$-shot episode, we construct a support set $\mathcal{S}$ and a query set $\mathcal{Q}$, both of which belong to the training set $\mathcal{D}_{\text{train}}$. The support set $\mathcal{S}$ consists of $N$ action categories, with $K$ labeled videos per category, i.e., $\mathcal{S}=\{(v_i^s, y_i^s)\}_{i=1}^{N\times K}$, where $v_i^s$ denotes the $i$-th support video and $y_i^s\in\{1,2,\ldots,N\}$ denotes its corresponding label. These samples provide limited prior information for the current episode. The query set $\mathcal{Q}$ consists of additional videos from the same $N$ categories, typically containing $M$ videos per category, i.e., $\mathcal{Q}=\{(v_j^q, y_j^q)\}_{j=1}^{N\times M}$. The model must classify the query samples in $\mathcal{Q}$ using only the information conveyed by the support set $\mathcal{S}$.

Let $f_\theta$ denote the FSAR model with parameters $\theta$. Within each training episode, $\theta$ is optimized by minimizing the classification loss on the query set. The objective is to learn a feature embedding space or distance metric in which samples from the same class are close while those from different classes are well separated. At evaluation time, performance is measured on a sequence of test episodes sampled from $\mathcal{D}_{\text{test}}$, whose classes are unseen during training; the average classification accuracy across these episodes serves as the primary metric of generalization.

\subsection{Overall framework}
\label{subsec3.2}

To tackle the challenges of temporal modeling, visual similarity, and the modality gap in few-shot action recognition, we propose a cross-modal learning-based framework termed CLIP-SPM. The overall pipeline is illustrated in Fig.~\ref{fig1}. The central idea is to learn video representation that are both highly discriminative via three complementary components---Hierarchical Synergistic Motion Refinement, Semantic Prototype Modulation, and Prototype--Anchor Dual Modulation---thereby yielding more discriminative class prototypes.

Given an $N$-way $K$-shot task, the support and query samples are denoted by $S = \{s^1, s^2,\ldots,s^{N\times K}\}$ and $Q = \{q^1,q^2,\ldots,q^{N\times M}\}$, where the superscript indexes videos. To reduce the computational cost, sparse sampling strategy~\cite{wang2016temporal} is applied to extract $T$ frames per video: $s^i = \{s^i_1, s^i_2, \ldots, s^i_T\}$ and $q^j = \{q^j_1, q^j_2, \ldots, q^j_T\}$. Then, the CLIP image encoder $\mathcal{V}$ extracts visual features. Specifically:
\begin{align}
  f_{s^i} = \mathcal{V}(s^i) \nonumber \\
  f_{q^j} = \mathcal{V}(q^j)
\end{align}
where $f_{s^i} \in \mathbb{R}^{T\times C}$ and $f_{q^j} \in \mathbb{R}^{T\times C}$ denote the visual features of the support and query videos, respectively, and $C$ is the number of feature channel. To alleviate linguistic polysemy, we adopt a prompt-ensemble strategy~\cite{wang2025actionclip}: multiple natural-language prompts of base classes are fed into the CLIP text encoder $\mathcal{T}$ to obtain text embeddings, which are then aggregated to produce a robust textual prompt representation. Specifically:
\begin{align}
  f_{tp_n^r} = \mathcal{T}(tp_n^r) \nonumber \\
  f_{tp_n} = \sum_{r=1}^R f_{tp_n^r}
\end{align}
where $tp_n^r$ denotes the textual prompt formed by combining the $r$-th ($1 \le r \le R$) template with the $n$-th ($1 \le n \le N$) base class, where $R$ is the number of prompt templates. The fused textual prompt features for the $n$-th class are denoted by $f_{tp_n}$. All prompt templates are listed in~\ref{app1}.

To address the temporal modeling problem, we design a Hierarchical Synergistic Motion Refinement (HSMR) module to exploit implicit, consistent motion information. The HSMR produces shallow and deep motion features from the frame tokens, and imposes a consistency constraint between them. Specifically:
\begin{align}
    m^{s}_{v},\, m^{d}_{v} &= \mathcal{HSMR}(f_{v}) \nonumber \\
    d_{h}(v) &= ConDis(m^{s}_{v},\, m^{d}_{v})
\end{align}
where $f_{v} \in f_S \cup f_Q$ with $f_S = \{f_{s^1}, f_{s^2}, \ldots, f_{s^{NK}}\}$ and $f_Q = \{f_{q^1}, f_{q^2}, \ldots, f_{q^{NM}}\}$. $m^{s}_{v}$ and $m^{d}_{v}$ denote shallow and deep motion features, respectively. The consistency distance $ConDis$ is described in Section~\ref{subsec4.2}.

To address the visual similarity challenge, we design a Semantic Fusion (SF) module that leverages motion and textual semantics to modulate discriminative visual features. Building on this, we propose the Semantic Prototype Modulation (SPM) strategy, which learns query-relevant prompts from textual prompts within the current episode. This reduces the modality gap caused by the absence of textual information in query samples. By modulating visual features with the learned prompts, SPM enhances the discriminability of features, especially for visually similar actions. Specifically:
\begin{align}
    f^{SPM}_{S},\; f^{SPM}_{Q} &= \mathcal{SPM}(f_S,\, f_Q,\, f_{TP}) \nonumber \\
    f^{SPM}_{\tilde{s}^n} &= \frac{1}{K}\sum_{k=1}^K f^{SPM}_{s^{(n-1)\times K+k}} \nonumber \\
    d_{spm}(f^{SPM}_{\tilde{s}^n},\, f^{SPM}_{q^j}) &= SeqDis(f^{SPM}_{\tilde{s}^n},\, f^{SPM}_{q^j})
\end{align}
where $ f_{TP} \in \mathbb{R}^{NK \times 1 \times C} $ represents the text prompt representation corresponding to the support samples in the current episode. $ f^{SPM}_{s^i}\in\mathbb{R}^{NK\times T\times C} $ and $ f^{SPM}_{q^j}\in\mathbb{R}^{NM\times T\times C} $ denote the semantic-modulated features for the support and query samples, respectively. $ f^{SPM}_{\tilde{s}^n}\in\mathbb{R}^{N\times T\times C} $ represents the class prototypes of $ f^{SPM}_{S} $. The sequence distance $ SeqDis $ is defined in Section~\ref{subsec4.2}.

In each episode, the support and query samples share a unified feature space, with the support samples providing ample discriminative information in this space. To fully leverage this property, we propose the Prototype-Anchor Dual Modulation (PADM) method. The PADM enhances the semantic consistency between query and support samples by introducing anchor features and modulates the support prototypes using the support samples to improve their discriminability. Specifically:
\begin{align}
    f^{PADM}_{S},\; f^{PADM}_{Q},\; f^{P}_{S},\; f^{A}_{Q} &= \mathcal{PADM}(f^{SPM}_{S},\, f^{SPM}_{Q}) \nonumber \\
    f^{PADM}_{\tilde{s}^n} &= \frac{1}{K}\sum_{k=1}^K f^{PADM}_{s^{(n-1)\times K+k}} \nonumber \\
    d_{padm}(f^{PADM}_{\tilde{s}^n}, f^{PADM}_{q^j}) &= SeqDis(f^{PADM}_{\tilde{s}^n},\, f^{PADM}_{q^j})  \nonumber \\
    & + SeqDis(f^{P}_{s^n},\, f^{A}_{q})
\end{align}
where $ f^{PADM}_{S} \in \mathbb{R}^{NK \times T \times C} $ and $ f^{PADM}_{Q} \in \mathbb{R}^{NM \times T \times C} $ represent the support and query features, respectively, modulated by class prototypes and a visual anchor. $ f^{P}_{S} \in \mathbb{R}^{N \times T \times C} $ denotes the modulated class prototypes, while $ f^{A}_{Q} \in \mathbb{R}^{1 \times T \times C} $ is the query-conditioned visual anchor. Finally, $ f^{PADM}_{\tilde{s}^n} \in \mathbb{R}^{N \times T \times C} $ represents the class prototypes of $ f^{PADM}_{S} $.

Finally, by combining the two sequence distances, a comprehensive distance between the support and query samples is computed. CLIP-SPM then performs few-shot classification based on this distance. The class probability distribution of a query video in each episode is derived from the support prototypes. Specifically:
\begin{align}
  d(v^{\tilde{s}}_n,v^q_j) & =\lambda_{1}d_{padm}(f^{PADM}_{\tilde{s}^n}, f^{PADM}_{q^j})  \nonumber \\ 
  &+\lambda_{2}d_{spm}(f^{SPM}_{\tilde{s}^n},\, f^{SPM}_{q^j}) \nonumber \\
  p(y^q_{j}=o\mid v^q_j,S) & =\frac{\exp\left(-d(v^{\tilde{s}}_o,v^q_j)\right)}{\sum_{n=1}^N\exp(-d(v^{\tilde{s}}_n,v^q_j))}
\end{align}
where $\lambda_1$ and $\lambda_2$ are the weights for the PADM and SPM sequence distances, respectively. $ v^{\tilde{s}}_n $ represents the class prototype for the $n$-th class. Next, we provide detailed descriptions of the three core components and their associated loss functions.

\subsection{Hierarchical synergistic motion regularization}
\label{subsec3.3}

In few-shot learning scenarios, relying solely on visual frame features (i.e., appearance features) is insufficient to fully capture the essence of actions. For example, the static frames of ``Running'' versus ``Jogging'' may appear visually similar, yet their motion patterns differ significantly. Motion features can effectively complement appearance features, enhancing the model's ability to distinguish between classes. However, existing FSAR methods~\cite{wang2023molo,guo2025consistency,deng2025clip} often fail to fully exploit the complementary relationship between motion and appearance features, underutilizing the potential of motion information. To address this, we design the Hierarchical Synergistic Motion Regularization (HSMR) module, which aligns shallow and deep motion features through consistency loss constraints to ensure the consistency of motion patterns across different layers.

Specifically, the HSMR module first extracts motion features through temporal differencing (referred to as shallow motion features), then modulates these motion features with appearance features, and finally extracts the deep motion features. The procedure is outlined as follows:
\begin{align}
m_{v}^{s} & =\mathcal{MFE}(f_{v}) \nonumber \\
m_{v}^{s},f_{v}^{s} & =S\mathcal{F}\left(m_{v}^{s},f_{v}\right) \nonumber \\
m_{v}^{d} & =\mathcal{MFE}(f_{v}^{s})
\end{align}
where $\mathcal{MFE}$ represents a motion feature extraction module, and $\mathcal{SF}$ denotes a semantic fusion module. $m_{v}^{s}$ and $m_{v}^{d}$ represent the shallow and deep motion features, respectively. The specific details of the $\mathcal{MFE}$ and $\mathcal{SF}$ will be provided in the following subsections.

\begin{figure}[t]
  \centering
  \includegraphics[width=\linewidth]{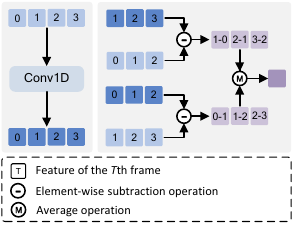}
  \caption{Illustration of the Motion Feature Extraction (MFE) module.}
  \label{fig2}
\end{figure}

\subsubsection{Motion feature extraction}

In few-shot action recognition, effectively capturing inter-frame motion information is crucial for distinguishing between actions. To this end, we adopt the Motion Feature Extraction (MFE) module, inspired by~\cite{deng2025clip}, which extracts motion features through differencing operations, as shown in Fig.~\ref{fig2}. The core idea behind the MFE is to explicitly capture motion information by applying forward and backward differencing between adjacent frames. Specifically, given the visual features of a video $f_{v} \in \mathbb{R}^{T\times C}$, the MFE computes the motion features as follows:
\begin{align}
m_t^f & =\Phi\left(f_{v}[t+1]\right)-f_{v}[t] \nonumber \\
m_t^b & =\Phi\left(f_{v}[t]\right)-f_{v}[t+1] \nonumber \\
m_{t} & =m_t^f+m_t^b \nonumber \\
m & =\sum_{t=1}^{T-1}m_t
\end{align}
where $\Phi$ represents a series of convolution operations, while $m_t^{f}$ and $m_t^{b}$ denote the forward and backward differencing features, respectively. The variable $m_t$ represents the local motion feature at time step $t$. This bidirectional differencing approach helps capture a more comprehensive context of motion, mitigating potential biases that may arise from unidirectional differencing. Finally, $m$ aggregates all short-term motion variations within the segment, serving as an effective motion descriptor.

\subsubsection{Semantic fusion}

\begin{figure*}[t]
  \centering
  \includegraphics[width=0.7\linewidth]{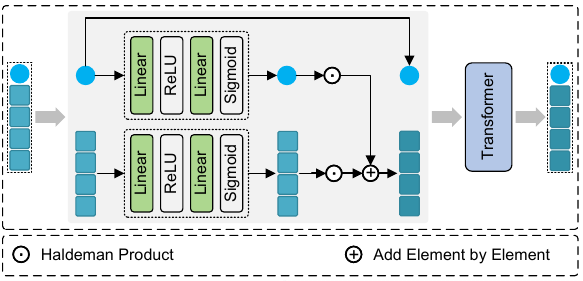}
  \caption{Architecture of the Semantic Fusion (SF) module.}
  \label{fig3}
\end{figure*}

In few-shot action recognition, effectively modeling the relationship between appearance and motion information is crucial. To this end, the MFE module first extracts motion features from consecutive video frames. Building on this, the Semantic Fusion (SF) module treats motion information as prompt information to modulate the appearance features. As shown in Fig.~\ref{fig3}, the SF module introduces a gating fusion mechanism that enables the adaptive integration of appearance features and motion prompts. This mechanism dynamically adjusts the contributions of appearance features and motion prompts through learnable gating weights. The gating weights are generated based on the contextual information of the current appearance and prompt features, determining which parts of the feature vectors should emphasize motion variation and which should retain more appearance details. This dynamic weighting strategy effectively mitigates potential redundancy or conflict issues that could arise from simple concatenation or linear fusion, thereby improving fusion efficiency in complex dynamic scenarios. Formally, the SF operation is defined as:
\begin{align}
f_v &= \Psi_v(f_{v}) \odot f_{v} + \Psi_{tp}(f_{tp}) \odot f_{tp} \nonumber \\
f_{tp},\, f_{v} &= Transformer(Concat(f_{tp},\, f_{v}))
\end{align}
where $\Psi_v$ and $\Psi_{tp}$ represent the visual gating network and the prompt gating network, respectively. $f_v$ and $f_{tp}$ denote the visual features and prompt features (e.g., motion or textual prompts), respectively. \textit{Transformer} refers to the Transformer encoder, \textit{Concat} denotes the feature concatenation operation, and $\odot$ represents the Hadamard product.

\subsection{Semantic prototype modulation}
\label{subsec3.4}

Existing FSAR methods~\cite{wang2024clip, ni2024multimodal} typically construct support prototypes by integrating visual and textual features, creating rich cross-modal semantic representation. However, query samples lack corresponding textual semantic support, leading to a significant modality gap between the support and query features. CLIP-CPM$^2$C~\cite{guo2025consistency} attempts to address this by introducing randomly generated query text embeddings and employing consistency loss constraints to learn query text representation. However, the randomly generated embeddings often show weak correlation with the corresponding query visual information, which can cause the optimization process to get stuck in local optima, making it difficult to close the modality gap. To address this issue, we propose the Semantic Prototype Modulation (SPM) strategy, which dynamically learns text prompt embeddings that are closely aligned with the visual features of query samples from the current episode, effectively bridging the modality gap. Then, the SPM integrates visual and textual prompt features through the SF module, modulating discriminative visual features with text semantics to tackle the visual similarity challenge. The process begins by learning the text prompt embeddings for both the support and query samples:
\begin{align} 
  \bar{f}_{TP_{S}} = \mathcal{PG}(f_{TP}, f_{S}) \nonumber \\
  \bar{f}_{TP_{Q}} = \mathcal{PG}(f_{TP}, f_{Q})
\end{align}
where $\mathcal{PG}$ denotes the prompt generator, which is described in further detail in the following subsection. $ \bar{f}_{TP_{S}} \in \mathbb{R}^{NK \times 1 \times C} $ and $ \bar{f}_{TP_{Q}} \in \mathbb{R}^{NM \times 1 \times C} $ represent the learned prompt features for the support and query samples, respectively. Subsequently, SPM dynamically fuses the visual features and textual prompt features through the SF module to modulate the discriminative visual features. Specifically:
\begin{align}
f_{tp_{s^{i}}},f_{s^{i}}=\mathcal{SF}\left(f_{tp_{s^{i}}},f_{s^{i}}\right) \nonumber \\
f_{tp_{q^{j}}},f_{q^{j}}=\mathcal{SF}\left(f_{tp_{q^{j}}},f_{q^{j}}\right) \nonumber \\
\bar{f}_{tp_{s^{i}}},\bar{f}_{s^{i}}=\mathcal{SF}\left(\bar{f}_{tp_{s^{i}}},f_{s^{i}}\right) \nonumber \\
\bar{f}_{tp_{q^{j}}},\bar{f}_{q^{j}}=S\mathcal{F}\left(\bar{f}_{tp_{q^{j}}},f_{q^{j}}\right)
\end{align}
where $ f_{tp_{s^i}} $ and $ f_{tp_{q^j}} $ represent the original prompt features of the support and query samples, respectively, while $ \bar{f}_{tp_{s^i}} $ and $ \bar{f}_{tp_{q^j}} $ denote the learned prompt representations for the support and query samples, respectively. $ f_{s_i} $ and $ f_{q_j} $ represent the corresponding visual features. $ \bar{f}_{s_i} $ and $ \bar{f}_{q_j} $ are the modulated visual features obtained by integrating the learned prompt features $ \bar{f}_{tp_{s^i}} $ and $ \bar{f}_{tp_{q^j}} $. To further improve the consistency between the dynamic and ground-truth features, SPM introduces a multi-level consistency loss. Specifically:
\begin{align}
d_s^{tp}(v^{s}_{i}) & =ConDis(f_{tp_{s^{i}}},\bar{f}_{tp_{s^{i}}}) \nonumber \\
d_s^{tp}(v^{q}_{j}) & =ConDis(f_{tp_{q^{j}}},\bar{f}_{tp_{q^{j}}}) \nonumber \\
d_{s}^{v}(v^{s}_{i}) & =ConDis\left(Concat(f_{tp_{s^i}},f_{s^i}),Concat(\bar{f}_{tp_{s^i}},\bar{f}_{s^i})\right) \nonumber \\
d_{s}^{v}(v^{q}_{j}) & =ConDis\left(Concat(f_{tp_{q^j}},f_{q^{j}}),Concat(\bar{f}_{tp_{q^{j}}},\bar{f}_{q^{j}})\right) \nonumber \\
d_{s}(v^{s}_{i}) & =d_{s}^{tp}(v^{s}_{i})+d_{s}^{v}(v^{s}_{i}) \nonumber \\
d_s(v^{q}_{j}) & =d_{s}^{tp}(v^{q}_{j})+d_{s}^{v}(v^{q}_{j})
\end{align}
where $d_s$ enforces consistency between the learned and real prompt features. The SPM effectively bridges the modality gap between the support and query sets by using a textual semantics--guided dynamic prompt generation mechanism and structured consistency constraints. Furthermore, SPM enhances the discriminability of visual features by modulating them with textual features.

\begin{figure}[t]
  \centering
  \includegraphics[width=0.8\linewidth]{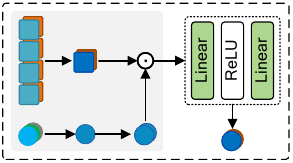}
  \caption{Illustration of the Prompt Generator (PG).}
  \label{fig4}
\end{figure}

\subsubsection{Prompt generator}

During the meta-testing phase, CLIP-FSAR~\cite{wang2024clip} typically relies solely on matching the visual features of query samples with support prototypes, as the true labels of query samples are unavailable. This approach inevitably introduces a modality gap between the support prototypes and query samples: the support prototypes contain rich textual semantics, while the query samples consist only of visual information. To address this issue, we adopt the prompt generator (PG) module~\cite{deng2025clip} (as shown in Fig.~\ref{fig4}). The PG fuses prompt embeddings from the support set with the query samples to generate query-relevant prompt embeddings, effectively substituting for the real prompt embeddings. Subsequently, the SPM module modulates the query samples using the learned prompt embeddings via the SF module, compensating for the missing textual information and bridging the modality gap. Specifically, given the textual prompt features $ f_{TP} \in \mathbb{R}^{NK \times 1 \times C} $ and visual features $ f_{Q} \in \mathbb{R}^{NM \times T \times C} $ of the query samples, the PG generates the query-relevant prompt representation as:
\begin{equation}
  \mathcal{P}\mathcal{G}\left(f_{TP},f_{Q}\right)=\Upsilon\left(\left(\frac{1}{NK}\sum_{o=1}^{NK}f_{tp_{o}}\right)\odot\left(\frac{1}{T}\sum_{t=1}^{T}f_{Q}[t]\right)\right)
\end{equation}
where $\Upsilon$ denotes a series of linear layers.

\subsection{Prototype-anchor dual modulation}
\label{subsec3.5}

In few-shot action recognition, the quality of support prototypes directly impacts the model's classification performance on query samples. Some methods~\cite{wang2024clip, guo2025consistency} typically construct class prototypes by simply averaging the support samples. However, such prototypes often fail to capture the underlying intra-class structure and key distinguishing features, which limits their representational capacity. To enhance the expressiveness of prototype features, we introduce the Prototype-Anchor Dual Modulation (PADM) concept, which dynamically refines both the support and query samples using class prototype features and anchor features. Specifically, class prototype features are used to refine the representation of the support samples, enhancing their discriminability. Anchor features, on the other hand, modulate the query samples, improving their semantic consistency with the global anchor features. Finally, by calculating the sequence distance between the support and query sample features, obtained through dual modulation, more accurate action category recognition can be achieved. The formal mathematical definition of this concept is as follows:
\begin{align}
  \tilde{f}_{S}&=Reshape(f_{S}) \nonumber \\
  f_{S}^{P}&=Mean(Concat(\tilde{f}_{S},\bar{f}_{Q}))  \nonumber \\
  f^{A}&=Mean(f_{S}^{P}) \nonumber \\
  f_{S}^{P},f_{S}&=Transformer(Concat(f_{S}^{P},f_{S}))  \nonumber \\
  f_{Q}^{A},\bar{f}_{Q}&=Transformer(Concat(f^{A}, \bar{f}_{Q}))
\end{align}
where $ \tilde{f}_S \in \mathbb{R}^{N \times K \times T \times C} $ and $ \bar{f}_Q \in \mathbb{R}^{1 \times NM \times T \times C} $ represent the visual features of the support and query samples, respectively. $ f_S^{P} \in \mathbb{R}^{N \times T \times C} $ denotes the class prototypes after modulation using the support features, and $ f^{A} \in \mathbb{R}^{1 \times T \times C} $ represents the anchor features. \textit{Reshape} refers to the operation of altering tensor dimensions, while \textit{Mean} denotes the averaging operation. Finally, the PADM module outputs four sets of features: the support features $ f_S $, the query features $ \bar{f}_Q $, the modulated class prototype $ f_S^{P} $, and the modulated query anchor $ f_Q^{A} $, all of which are used for subsequent computations.

\subsection{Loss function}
\label{subsec3.6}

To optimize both classification performance and generalization ability, CLIP-SPM is trained using a multi-task loss function that combines classification loss and consistency loss. The overall definition of the loss function is as follows:
\begin{align}
\mathcal{L}_{CE}&=-\frac{1}{NM}\sum_{j =1}^{N M}\log p(y^{q}_{j}|v^{q}_{j},S) \nonumber \\
\mathcal{L}_{H}&=\sum_{f_{v}\in f_{S}\cup f_{Q}}d _{h}(f_{v}) \nonumber \\
\mathcal{L}_{S}&=\sum_{v^{s}_{i}\in S}d_{s}(v^{s}_{i})+ \sum_{v^{q}_{j}\in Q}d_{s}(v^{q}_{j}) \nonumber \\
\mathcal{L}&=\mathcal{L}_{CE}+\lambda_{3}\mathcal{L }_{H}+\lambda_{4}\mathcal{L}_{S}
\end{align}
where $\mathcal{L}_{CE}$ is the standard cross-entropy classification loss, which aims to minimize the negative log-likelihood of the true label $y^{q}_{j}$ for the query sample $v^{q}_{j}$. This term constitutes the main loss function and directly optimizes the model's classification accuracy. Both $\mathcal{L}_H$ and $\mathcal{L}_S$ are applied to all visual samples from the support and query sets (i.e., $S \cup Q$). $\mathcal{L}_{H}$ aims to improve the consistency between shallow and deep motion features, while $\mathcal{L}_{S}$ seeks to enhance the consistency between the learned prompt representation and the original prompt features.

The final overall loss $\mathcal{L}$ is the weighted sum of these three terms, with hyperparameters $\lambda_3$ and $\lambda_4$ used to balance the importance between the classification task and the two consistency constraints. By jointly optimizing this multi-objective loss function, CLIP-SPM achieves excellent classification performance while maintaining strong generalization ability.

\section{Experiments}\label{sec4}

\subsection{Datasets}\label{subsec4.1}

We evaluate our model on five widely used few-shot action recognition benchmarks: Kinetics~\cite{carreira2017quo}, SSv2-Full~\cite{goyal2017something}, SSv2-Small~\cite{goyal2017something}, UCF101~\cite{soomro2012ucf101}, and HMDB51~\cite{kuehne2011hmdb}. Dataset partitions follow standard protocols~\cite{cao2020few, zhu2018compound, zhu2020label, zhang2020few} to ensure a fair comparison. For Kinetics-400 and SSv2-Small, we adopt the CMN~\cite{zhu2018compound} and CMN-J~\cite{zhu2020label} splits, where a 100-class subset is randomly selected, with 64/12/24 classes for training/validation/testing, and 100 samples per class. SSv2-Full follows the OTAM~\cite{cao2020few} protocol. Unlike SSv2-Small, it uses all available samples per class, containing 77,500 training videos, 1,926 validation videos, and 2,854 test videos. For UCF101 and HMDB51, we follow the ARN~\cite{zhang2020few} splits. UCF101 includes 70/10/21 classes for training/validation/testing, with 9,154/1,421/2,745 videos, respectively. HMDB51 includes 31/10/10 classes, with 4,280/1,194/1,292 videos, respectively.

\subsection{Experimental settings}\label{subsec4.2}

Following the approach of TSN~\cite{wang2016temporal}, we uniformly sample $T = 8$ frames from each video. During training, standard data augmentation techniques are applied, including random horizontal flipping, and random $224 \times 224$ cropping. During inference, center cropping with $224 \times 224$ is applied. We adopt a standard 5-way $K$-shot episodic protocol, where $K \in \{1, 3, 5\}$, each episode contains 5 classes, with $K$ support samples and $M=4$ query samples per class. For HMDB51, UCF101, and Kinetics, we randomly sample 20,000 training episodes, while for SSv2-FULL and SSv2-Small, we sample 100,000 episodes to better account for their motion-centric characteristics. The vision-language backbone is instantiated with CLIP encoders using ResNet-50 and ViT-B/16. In all experiments, the visual encoder is fine-tuned, while the text encoder remains frozen. The network is optimized using the Adam optimizer~\cite{kingma2014adam} with a learning rate of $1 \times 10^{-5}$ and a weight decay of $5 \times 10^{-5}$. The loss weights are set as $\lambda_1 = 1$, $\lambda_2 = 0.5$, and $\lambda_3 = \lambda_4 = 0.001$. To stabilize training and reduce memory consumption, gradients are accumulated over 16 iterations before each optimizer step. For evaluation, we follow the same 5-way $K$-shot protocol and randomly sample 10,000 test episodes from the test splits of each dataset, reporting the mean classification accuracy. In line with common practice, we compute the consistency distance using Euclidean distance~\cite{liberti2014euclidean} and the sequence distance using OTAM~\cite{cao2020few}, ensuring fair comparison with prior work. All experiments are implemented using PyTorch 2.4.1 and conducted on NVIDIA A100 GPUs.

\begin{table*}[!tb]
\centering
\footnotesize
\caption{Comparison with prior few-shot action recognition methods on HMDB51 and UCF101 under 1/3/5-shot settings. All numbers are Top-1 accuracy (\%). The ``Pre-training'' column specifies the backbone and its pretraining source (INet-RN50 = ResNet-50 pretrained on ImageNet; CLIP-RN50 / CLIP-ViT-B = CLIP encoders). Best results are in \textbf{bold}, while second-best results are in \textit{italics}. ``-'' indicates that the metric was not reported.}
\label{tab1}
\begin{tabular}{l l l | c c c | c c c}
\toprule
Method & Reference & Pre-training 
& \multicolumn{3}{c|}{HMDB51} 
& \multicolumn{3}{c}{UCF101} \\
& & & 1-shot & 3-shot & 5-shot & 1-shot & 3-shot & 5-shot \\
\midrule
OTAM~\cite{cao2020few}                    & CVPR'20   & INet-RN50  & 54.5            & -               & 66.1            & 79.9             & -               & 88.9            \\
TRX~\cite{perrett2021temporal}            & CVPR'21   & INet-RN50  & 52.9            & -               & 75.6            & 77.3             & -               & 96.1            \\
STRM~\cite{thatipelli2022spatio}          & CVPR'22   & INet-RN50  & 54.1            & -               & 77.3            & 79.2             & -               & 96.9            \\
HyRSM~\cite{wang2022hybrid}               & CVPR'22   & INet-RN50  & 60.3            & 71.7            & 76.0            & 83.9             & 93.0            & 94.7            \\
MTFAN~\cite{wu2022motion}                 & CVPR'22   & INet-RN50  & 59.0            & -               & 74.6            & 84.8             & -               & 95.1            \\
TA2N~\cite{li2022ta2n}                    & AAAI'22   & INet-RN50  & 59.7            & -               & 73.9            & 81.9             & -               & 95.1            \\
MoLo~\cite{wang2023molo}                  & CVPR'23   & INet-RN50  & 60.8            & 72.0            & 77.4            & 86.0             & 93.5            & 95.5            \\
DPN~\cite{jiang2024dual}                  & NEUCOM'24 & INet-RN50  & 60.2            & -               & 76.6            & 87.7             & -               & 98.0            \\
HCR~\cite{li2024hierarchical}             & CVIU'24   & INet-RN50  & 48.6            & -               & 63.8            & 71.8             & -               & 87.3            \\
TEAM~\cite{lee2025temporal}               & CVPR'25   & INet-RN50  & 62.8            & -               & 78.4            & 87.2             & -               & 96.2            \\
\midrule
CLIP-FSAR~\cite{wang2024clip}             & IJCV'23   & CLIP-RN50  & 69.2            & 77.6            & 80.3            & 91.3             & 95.1            & 97.0            \\
CLIP-FSAR~\cite{wang2024clip}             & IJCV'23   & CLIP-ViT-B & 75.8            & 84.1            & 87.7            & \underline{96.6} & \textbf{98.4}   & \textit{99.0} \\
STEN~\cite{chen2024cross}                 & APIN'24   & CLIP-RN50  & 59.1            & -               & 78.1            & 83.5             & -               & 96.8            \\
STEN~\cite{chen2024cross}                 & APIN'24   & CLIP-ViT-B & 69.9            & -               & 81.6            & 91.5             & -               & 96.8            \\
CLIP-M$^{2}$DF~\cite{guo2024multi}        & KBS'24    & CLIP-RN50  & 66.8            & 79.4            & 83.0            & \textbf{94.3}    & \textbf{98.0}   & \textbf{98.8}   \\
CLIP-M$^{2}$DF~\cite{guo2024multi}        & KBS'24    & CLIP-ViT-B & \textit{77.0} & 84.1            & 88.0            & \textbf{97.0}    & 98.1            & \textbf{99.3}   \\
CLIP-CPM$^{2}$C~\cite{guo2025consistency} & NEUCOM'25 & CLIP-RN50  & 66.3            & 78.4            & 81.2            & 91.1             & 96.2            & 97.4            \\
CLIP-CPM$^{2}$C~\cite{guo2025consistency} & NEUCOM'25 & CLIP-ViT-B & 75.9            & 84.4            & 88.0            & 95.0             & 98.1            & 98.6            \\
CLIP-MEI~\cite{deng2025clip}              & KBS'25    & CLIP-RN50  & \textit{76.4} & \textbf{83.0}   & \textbf{84.9}   & 93.7             & 97.0            & 97.5            \\
CLIP-MEI~\cite{deng2025clip}              & KBS'25    & CLIP-ViT-B & 75.9            & \textit{85.9} & \textit{88.5} & 94.9             & \textit{98.2} & 98.9            \\
\midrule
\rowcolor{gray!10}
CLIP-SPM                                  & -         & CLIP-RN50  & \textbf{77.4}   & \textit{82.6} & \textit{84.5} & \textit{93.8}  & \textit{97.2} & \textit{97.7} \\
\rowcolor{gray!10}
CLIP-SPM                                  & -         & CLIP-ViT-B & \textbf{78.2}   & \textbf{86.3}   & \textbf{88.6}   & 96.2             & \textit{98.2} & 98.7            \\
\bottomrule
\end{tabular}
\end{table*}

\begin{table*}[!tb]
\centering
\footnotesize
\caption{Comparison with prior few-shot action recognition methods on Kinetics, SSv2-Full, and SSv2-Small under 1/3/5-shot settings. All numbers are Top-1 accuracy (\%). The ``Pre-training'' column specifies the backbone and its pretraining source (INet-RN50 = ResNet-50 pretrained on ImageNet; CLIP-RN50 / CLIP-ViT-B = CLIP encoders). Best results are in \textbf{bold}, while second-best results are in \textit{italics}. ``-'' indicates that the metric was not reported. ``*'' stands for the results of our implementation.}
\label{tab2}
\begin{tabular}{l l l | c c c | c c c | c c c}
\toprule
Method & Reference & Pre-training & \multicolumn{3}{c|}{Kinetics} & \multicolumn{3}{c|}{SSv2-FULL} & \multicolumn{3}{c}{SSv2-Small} \\
& & & 1-shot & 3-shot & 5-shot & 1-shot & 3-shot & 5-shot & 1-shot & 3-shot & 5-shot \\
\midrule
OTAM~\cite{cao2020few}                     & CVPR'20   & INet-RN50  & 73.0            & -               & 85.5            & 42.8             & -               & 52.3            & 38.9            & -               & 48.1            \\
TRX~\cite{perrett2021temporal}             & CVPR'21   & INet-RN50  & 63.6            & -               & 85.9            & 42.0             & -               & 64.6            & 36.0            & -               & 59.4            \\
STRM~\cite{thatipelli2022spatio}           & CVPR'22   & INet-RN50  & 65.3            & -               & 86.7            & 42.9             & -               & 68.1            & -               & -               & -               \\
HyRSM~\cite{wang2022hybrid}                & CVPR'22   & INet-RN50  & 73.7            & 83.5            & 86.1            & 54.3             & 65.1            & 69.0            & 40.6            & 52.3            & 56.1            \\
MTFAN~\cite{wu2022motion}                  & CVPR'22   & INet-RN50  & 74.6            & -               & 87.4            & 45.7             & -               & 60.4            & -               & -               & -               \\
TA2N~\cite{li2022ta2n}                     & AAAI'22   & INet-RN50  & 72.8            & -               & 85.8            & 47.6             & -               & 61.0            & -               & -               & -               \\
MoLo~\cite{wang2023molo}                   & CVPR'23   & INet-RN50  & 74.0            & 83.7            & 85.6            & 56.6             & 67.0            & 70.6            & 42.7            & 52.9            & 56.4            \\
DPN~\cite{jiang2024dual}                   & NEUCOM'24 & INet-RN50  & 76.3            & -               & 90.7            & 46.8             & -               & 64.7            & -               & -               & -               \\
HCR~\cite{li2024hierarchical}              & CVIU'24   & INet-RN50  & 70.2            & -               & 81.4            & -                & -               & -               & -               & -               & -               \\
TEAM~\cite{lee2025temporal}                & CVPR'25   & INet-RN50  & 75.1            & -               & 88.2            & -                & -               & -               & -               & -               & -               \\
\midrule
CLIP-FSAR~\cite{wang2024clip}              & IJCV'23   & CLIP-RN50  & 87.6            & 90.7            & 91.9            & 58.1             & 60.7            & 62.8            & \textbf{52.0}   & 54.0            & 55.8            \\
CLIP-FSAR~\cite{wang2024clip}              & IJCV'23   & CLIP-ViT-B & 89.7            & 94.2            & 95.0            & 61.9             & 68.1            & 72.1            & \textit{54.5} & 58.6            & 61.8            \\
STEN~\cite{chen2024cross}                  & APIN'24   & CLIP-RN50  & 67.0            & -               & 85.4            & -                & -               & -               & 36.4            & -               & 57.3            \\
STEN~\cite{chen2024cross}                  & APIN'24   & CLIP-ViT-B & 83.0            & -               & 91.4            & -                & -               & -               & 38.9            & -               & 58.2            \\
CLIP-M$^{2}$DF~\cite{guo2024multi}         & KBS'24    & CLIP-RN50  & \textit{90.1} & \textit{92.0} & \textit{93.5} & 56.9             & 63.9            & 68.0            & \textit{51.6} & \textbf{56.3}   & \textbf{60.4}   \\
CLIP-M$^{2}$DF~\cite{guo2024multi}         & KBS'24    & CLIP-ViT-B & 90.9            & \textit{95.1} & \textbf{96.2}            & 60.1             & 68.9            & 72.7            & 52.0            & 58.9            & \textit{62.8} \\
CLIP-CPM$^{2}$C~\cite{guo2025consistency}  & NEUCOM'25 & CLIP-RN50  & 88.6            & 91.5            & 93.1            & 58.0             & 61.8            & 64.0            & 51.5            & 54.8            & 57.1            \\
CLIP-CPM$^{2}$C~\cite{guo2025consistency}  & NEUCOM'25 & CLIP-ViT-B & \textit{91.0} & \textit{95.1} & 95.5            & 60.1             & 68.9            & 72.8            & 52.3            & \textit{59.9} & 62.6            \\
CLIP-MEI~\cite{deng2025clip}               & KBS'25    & CLIP-RN50  & \textit{90.1} & \textbf{92.9}   & \textbf{93.6}   & \textit{61.6}$^*$ & \textit{65.9}$^*$ & \textit{68.5}$^*$ & -               & -               & -               \\
CLIP-MEI~\cite{deng2025clip}               & KBS'25    & CLIP-ViT-B & 87.7            & \textbf{95.3}   & \textit{96.0} & \textit{66.3}$^*$ & \textbf{75.9}$^*$   & \textbf{79.0}$^*$   & -               & -               & -               \\
\midrule
\rowcolor{gray!10}
CLIP-SPM                                   & -         & CLIP-RN50  & \textbf{90.9}   & \textbf{92.9}   & \textit{93.5} & \textbf{62.3}    & \textbf{67.0}   & \textbf{69.4}   & 50.8            & \textit{54.9} & \textit{58.5} \\
\rowcolor{gray!10}
CLIP-SPM                                   & -         & CLIP-ViT-B & \textbf{92.8}   & 94.1            & 94.3            & \textbf{66.7}    & \textit{74.8} & \textit{77.3} & \textbf{57.8}   & \textbf{62.4}   & \textbf{66.2}   \\
\bottomrule
\end{tabular}
\end{table*}

\subsection{Quantitative results}\label{subsec4.3}

We compare CLIP-SPM with representative few-shot action recognition approaches across five standard benchmarks: HMDB51, UCF101, Kinetics, SSv2-Full, and SSv2-Small, under the 1/3/5-shot evaluation protocol. The results with CLIP-RN50 and CLIP-ViT-B backbones are summarized in Tables~\ref{tab1} and \ref{tab2}. Overall, CLIP-SPM achieves strong and consistently competitive performance across all datasets, with particularly notable gains in the most data-scarce settings and on motion-centric benchmarks.

On HMDB51 and UCF101 (Table~\ref{tab1}), CLIP-SPM with the CLIP-ViT-B backbone reaches 78.2\% / 86.3\% / 88.6\% and 96.2\% / 98.2\% / 98.7\% in the 1/3/5-shot settings, respectively, surpassing or matching the strongest prior CLIP-based models. Even with CLIP-RN50, CLIP-SPM attains the highest reported 1-shot accuracy on HMDB51 and remains competitive at 3-shot and 5-shot, indicating that the proposed semantic prototype modulation mechanisms substantially enhance visual discriminability beyond what is achievable with the backbone alone.

On Kinetics, SSv2-Full, and SSv2-Small (Table~\ref{tab2}), which emphasize temporal reasoning and fine-grained motion cues, the gains are even more pronounced. CLIP-SPM achieves the strongest 1-shot performance on Kinetics for both backbones (90.9\% for RN50 and 92.8\% for ViT-B), and sets new state-of-the-art RN50 results on SSv2-Full across all shot settings. On SSv2-Small, CLIP-SPM with CLIP-ViT-B delivers the highest accuracies among all compared methods, reaching 57.8\%, 62.4\%, and 66.2\% in the 1/3/5-shot settings and improving upon the previous best by up to +3.4 percentage points; the RN50 variant remains within a small margin of the strongest RN50 baselines.

These consistent improvements across appearance-centric (HMDB51, UCF101) and motion-centric (Kinetics, SSv2-Full, SSv2-Small) benchmarks can be attributed to the complementary roles of the three core components. HSMR provides temporally consistent motion cues, benefiting datasets with complex dynamics such as SSv2-Full and SSv2-Small. SPM dynamically generates query-relevant prompts to bridge the semantic gap between support prototypes (vision + text) and visual-only queries. By fusing these prompts with visual features, it enhances the discriminability between visually similar actions. PADM further refines support prototypes and aligns query features to a global anchor, improving cross-set consistency and sharpening class decision boundaries.

In summary, across five benchmarks and two CLIP backbones, CLIP-SPM consistently improves performance under all shot settings, with the largest absolute gains observed in the 1-shot regime and on datasets requiring fine-grained motion understanding. These results confirm that the proposed framework provides robust temporal modeling, enhanced cross-modal alignment, and more discriminative prototypes, collectively improving the performance of the model in few-shot action recognition.

\begin{table*}[tb]
\centering
\footnotesize
\caption{Component analysis. Best results are in \textbf{bold}.}
\label{tab3}
\setlength{\tabcolsep}{6pt}
\begin{tabular}{ccc|ccc|ccc}
\toprule
\multicolumn{3}{c|}{Component} &
\multicolumn{3}{c|}{HMDB51} &
\multicolumn{3}{c}{SSv2-Small} \\
\cmidrule(lr){1-3}\cmidrule(lr){4-6}\cmidrule(lr){7-9}
HSMR & SPM & PADM &
1-shot & 3-shot & 5-shot &
1-shot & 3-shot & 5-shot \\
\midrule
\xmark & \xmark & \xmark & 64.5 & 75.5 & 79.3 & 41.0 & 49.2 & 54.4 \\
\midrule
\cmark & \xmark & \xmark & 64.3 & 75.6 & 79.4 & 41.3 & 49.4 & 54.8 \\
\xmark & \cmark & \xmark & 74.0 & 79.9 & 81.9 & 50.0 & 52.3 & 56.0 \\
\xmark & \xmark & \cmark & 68.4 & 79.3 & 83.1 & 41.1 & 51.1 & 55.9 \\
\midrule
\cmark & \cmark & \xmark & 73.8 & 79.3 & 82.1 & 48.6 & 52.4 & 56.4 \\
\cmark & \xmark & \cmark & 68.9 & 80.3 & 83.8 & 40.7 & 51.6 & 57.1 \\
\xmark & \cmark & \cmark & 76.9 & 82.4 & 84.4 & 50.6 & 54.8 & 58.1 \\
\midrule
\cmark & \cmark & \cmark & \textbf{77.4} & \textbf{82.6} & \textbf{84.5} & \textbf{50.8} & \textbf{54.9} & \textbf{58.5} \\
\bottomrule
\end{tabular}
\end{table*}

\subsection{Ablation studies}\label{subsec4.4}

To systematically assess the contribution of each module and design choice within CLIP-SPM, we conduct a comprehensive set of ablation studies. Unless otherwise stated, all experiments use CLIP-RN50 as the default backbone and follow the standard 5-way $K$-shot evaluation protocol.

\subsubsection{Component analysis}

To evaluate the contribution of each component in CLIP-SPM, we conduct an ablation study on HMDB51 and SSv2-Small under 1-, 3-, and 5-shot settings (Table~\ref{tab3}). The baseline model---without temporal, semantic, or prototype modulation---achieves the lowest accuracies, underscoring the need for additional refinement mechanisms in few-shot action recognition. Introducing HSMR yields modest improvements by enhancing motion consistency through hierarchical motion regularization. However, as HSMR focuses solely on motion cues, it does not sufficiently mitigate visual similarity or the support-query modality discrepancy, suggesting the necessity of complementary semantic and prototype-level adjustments. In contrast, SPM produces markedly larger gains, especially in the 1-shot regime. By generating query-relevant prompts and integrating them with visual features, SPM substantially increases discriminability among fine-grained actions and effectively narrows the modality gap. These results highlight the central role of semantic modulation in improving cross-modal alignment and representation quality. The PADM component also contributes positively by refining support prototypes and aligning query features with an anchor representation. While its improvements are slightly less pronounced than those of SPM, PADM strengthens inter-class separation and enhances support-query consistency, confirming its complementary effect. Combinations of modules further validate their synergy. Pairing SPM with PADM yields notable improvements, demonstrating that jointly modulating prototypes and query features enhances representation stability and discriminability. Ultimately, the full CLIP-SPM model-integrating HSMR, SPM, and PADM-achieves the highest performance across all datasets and shot settings. This confirms that addressing temporal modeling, semantic alignment, and prototype refinement in a unified framework.

\subsubsection{Feature fusion strategies in the SF module}

We investigate the contribution of different feature fusion strategies within the Semantic Fusion (SF) module by comparing the roles of Gate, Sum, and Concat across HMDB51 and SSv2-Small under 1-shot, 3-shot, and 5-shot settings (Table~\ref{tab4}). Among these, Concat serves as a fundamental baseline, as it ensures that visual and prompt features are properly aligned prior to the global attention operation in the Transformer. Beyond this baseline, incorporating Gate and Sum further improves fusion quality. Gate adaptively regulates the contribution of each feature stream, while Sum provides an additional residual pathway that stabilizes feature integration. Their combined use yields more coherent and discriminative fused representations. The best performance is observed when Gate, Sum, and Concat are applied jointly, demonstrating that these strategies are mutually reinforcing. Together, they support more effective blending of visual and semantic cues, ultimately enhancing the model's ability to distinguish fine-grained actions under few-shot conditions.

\begin{table*}[tb]
\centering
\footnotesize
\caption{Ablation study of feature fusion strategies in the SF module. Best results are in \textbf{bold}.}
\label{tab4}
\setlength{\tabcolsep}{6pt}
\begin{tabular}{ccc|ccc|ccc}
\toprule
\multicolumn{3}{c|}{Feature Fusion Strategy} &
\multicolumn{3}{c|}{HMDB51} &
\multicolumn{3}{c}{SSv2-Small} \\
\cmidrule(lr){1-3}\cmidrule(lr){4-6}\cmidrule(lr){7-9}
Gate & Sum & Concat &
1-shot & 3-shot & 5-shot &
1-shot & 3-shot & 5-shot \\
\midrule
\xmark & \xmark & \cmark & 76.0 & 81.4 & 84.4 & 50.7 & 52.8 & 58.2 \\
\xmark & \cmark & \cmark & 76.2 & 82.4 & 84.4 & 50.3 & \textbf{55.4} & 58.2 \\
\cmark & \cmark & \cmark & \textbf{77.4} & \textbf{82.6} & \textbf{84.5} & \textbf{50.8} & 54.9 & \textbf{58.5} \\
\bottomrule
\end{tabular}
\end{table*}

\begin{table*}[tb] 
\centering
\footnotesize
\caption{Ablation study of the consistency constraints for support and query sample modulation in the SPM module. Best results are in \textbf{bold}.}
\label{tab5}
\begin{tabular}{c|ccc|ccc}
\toprule
\multirow{2}{*}{Configuration} & \multicolumn{3}{c|}{HMDB51} & \multicolumn{3}{c}{SSv2-Small} \\
\cmidrule(lr){2-4} \cmidrule(lr){5-7}
 & 1-shot & 3-shot & 5-shot & 1-shot & 3-shot & 5-shot \\
\midrule
No Consistency Constraint     & 75.7          & 80.7          & 83.0          & 50.7          & 54.4          & 57.7          \\
Only Support Consistency      & 76.7          & \textbf{83.7} & 84.2          & 50.1          & 54.9          & \textbf{58.9} \\
Only Query Consistency        & 77.0          & 82.4          & 84.0          & 49.4          & \textbf{55.1} & 58.7          \\
Support and Query Consistency & \textbf{77.4} & 82.6          & \textbf{84.5} & \textbf{50.8} & 54.9          & 58.5          \\
\bottomrule
\end{tabular}
\end{table*}

\subsubsection{Consistency constraints in the SPM module}

We examine the role of different consistency constraints in the SPM module by comparing four configurations: no consistency constraint, support-only consistency, query-only consistency, and the joint application of both. Table~\ref{tab5} reports results on HMDB51 and SSv2-Small under 1-, 3-, and 5-shot settings. The configuration without any consistency constraint yields the weakest performance, particularly in the 1-shot regime, underscoring the importance of regularizing the semantic alignment process. Introducing support-side consistency leads to noticeable gains---most prominently in the 3-shot and 5-shot settings---indicating that refining support features stabilizes class prototypes and improves downstream matching. Applying consistency only to query features yields modest improvements as well, especially in 1-shot and 3-shot scenarios, suggesting that query-side regularization contributes to reducing the modality gap. The combined support-query consistency achieves the strongest overall performance, producing the highest accuracies in the 1-shot and 5-shot settings on HMDB51 and in the 1-shot settings on SSv2-Small, while remaining competitive in the remaining configurations. These results indicate that jointly regularizing both support and query features provides more coherent alignment across the two sets. This dual constraint effectively strengthens semantic consistency and enhances the reliability of prototype-query matching, ultimately yielding more robust few-shot action recognition performance.

\begin{figure}[tb]
  \centering
  \begin{subfigure}[t]{0.48\linewidth}
    \centering
    \includegraphics[width=\linewidth]{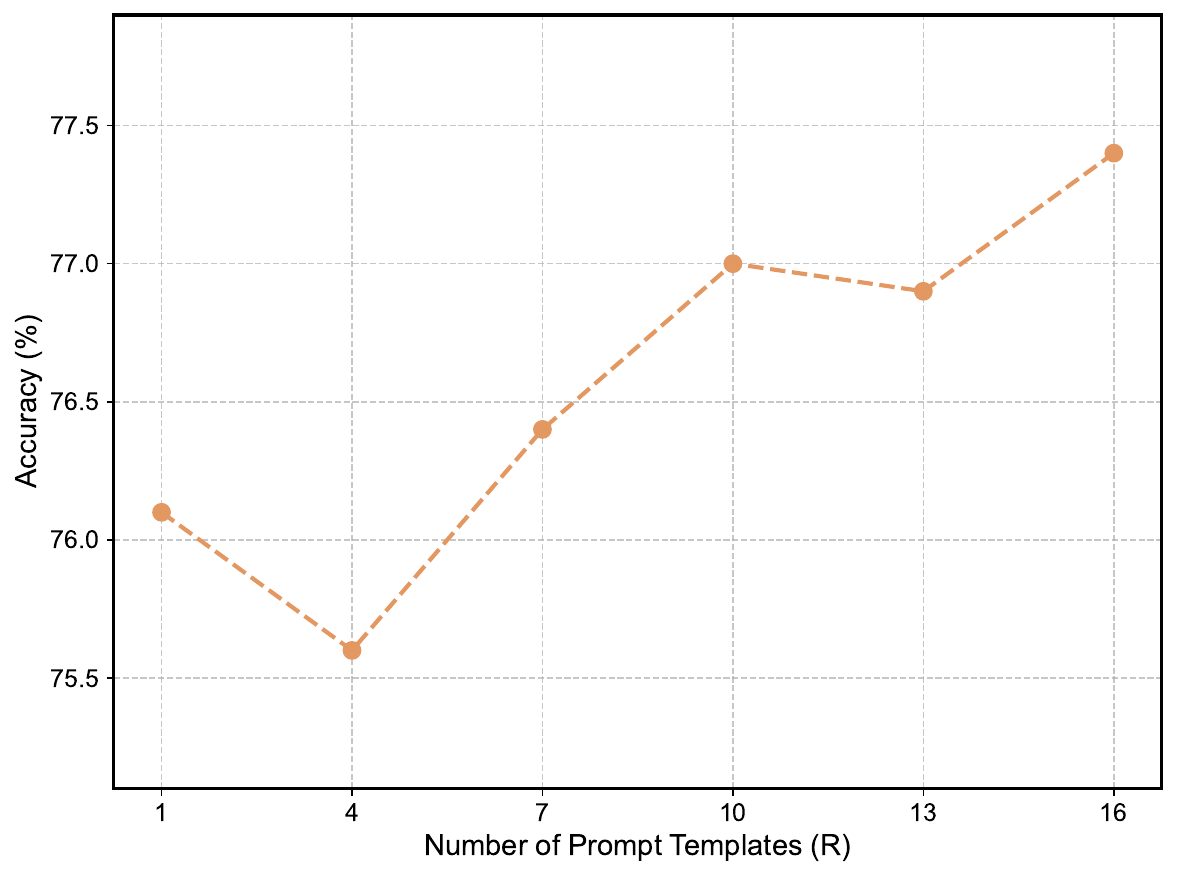}
    \subcaption{HMDB51}
    \label{fig5a}
  \end{subfigure}
  \hfill
  \begin{subfigure}[t]{0.48\linewidth}
    \centering
    \includegraphics[width=\linewidth]{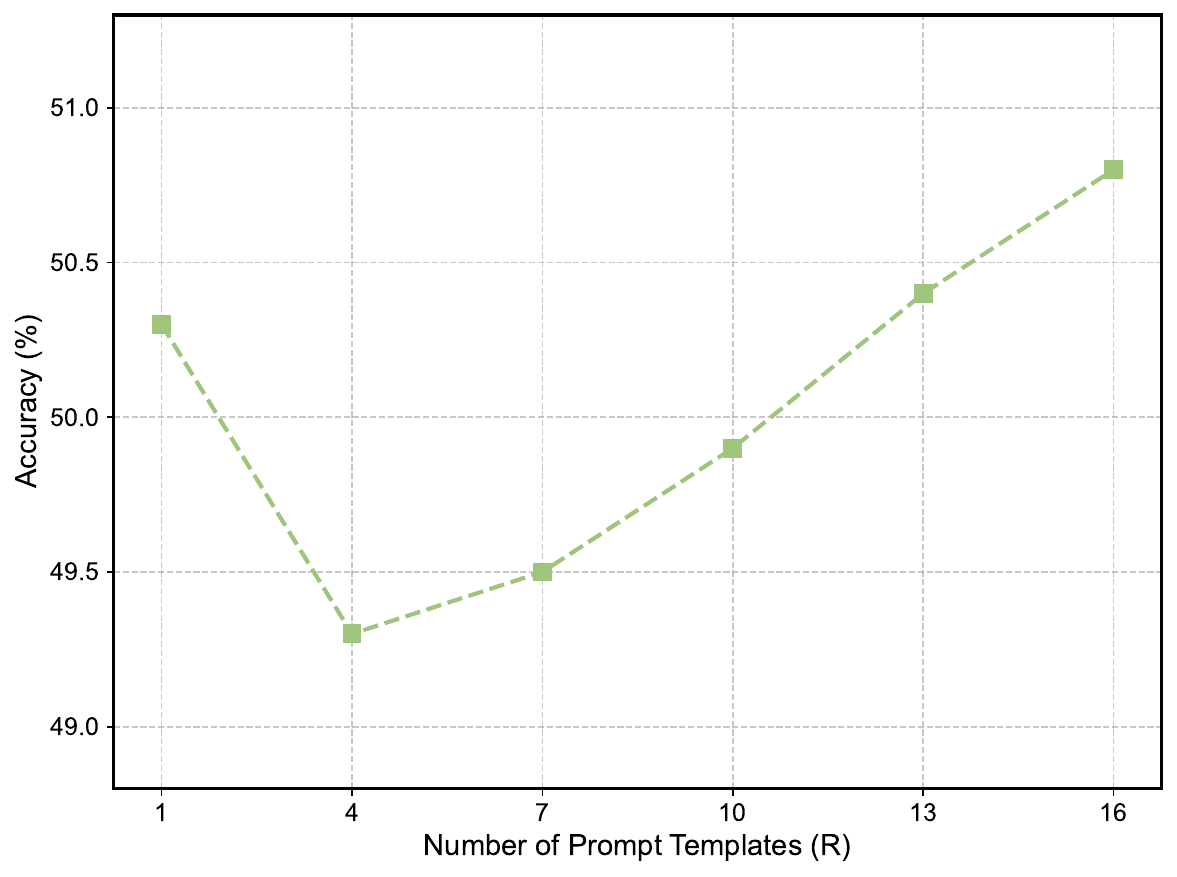}
    \subcaption{SSv2-Small}
    \label{fig5b}
  \end{subfigure}

  \caption{Ablation study of the number of prompt templates $R$ on HMDB51 and SSv2-Small under the 5-way 1-shot setting.}
  \label{fig5}
\end{figure}

\begin{figure*}[tb]
  \centering

  \begin{subfigure}[t]{0.3\linewidth}
    \centering
    \includegraphics[width=\linewidth]{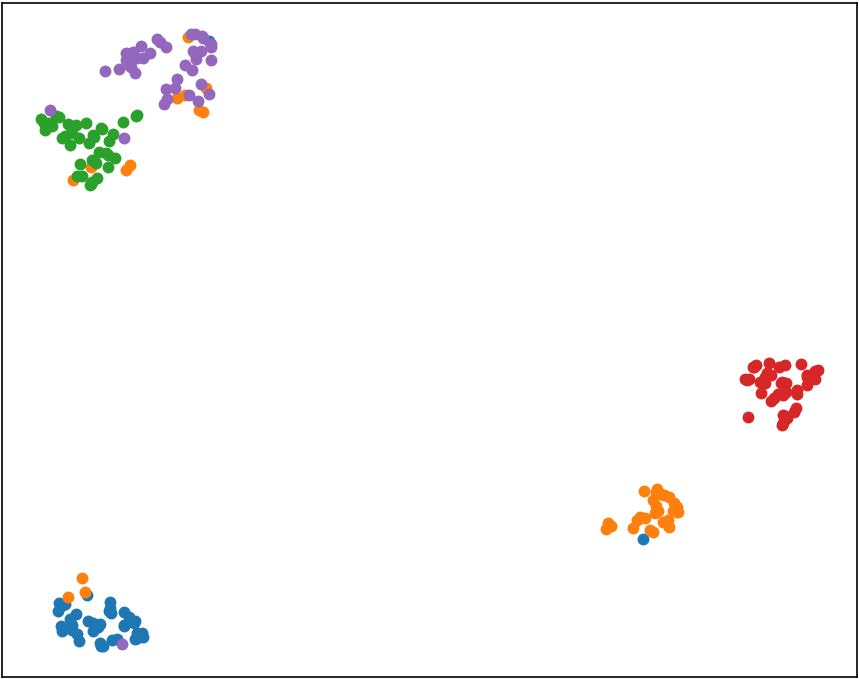}
    \subcaption{HMDB51}
    \label{fig6a}
  \end{subfigure}
  \hfill
  \begin{subfigure}[t]{0.3\linewidth}
    \centering
    \includegraphics[width=\linewidth]{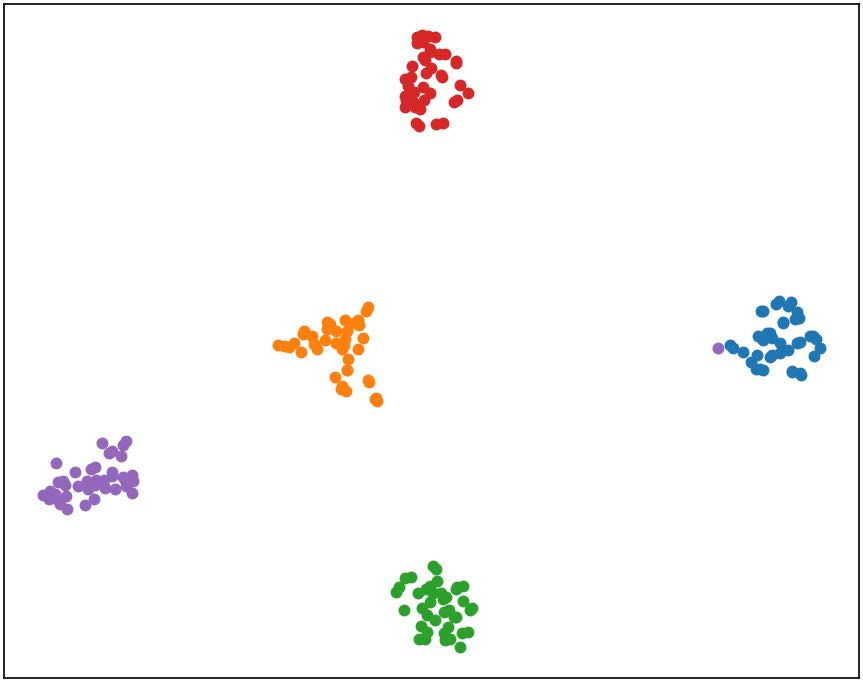}
    \subcaption{UCF101}
    \label{fig6b}
  \end{subfigure}
  \hfill
  \begin{subfigure}[t]{0.3\linewidth}
    \centering
    \includegraphics[width=\linewidth]{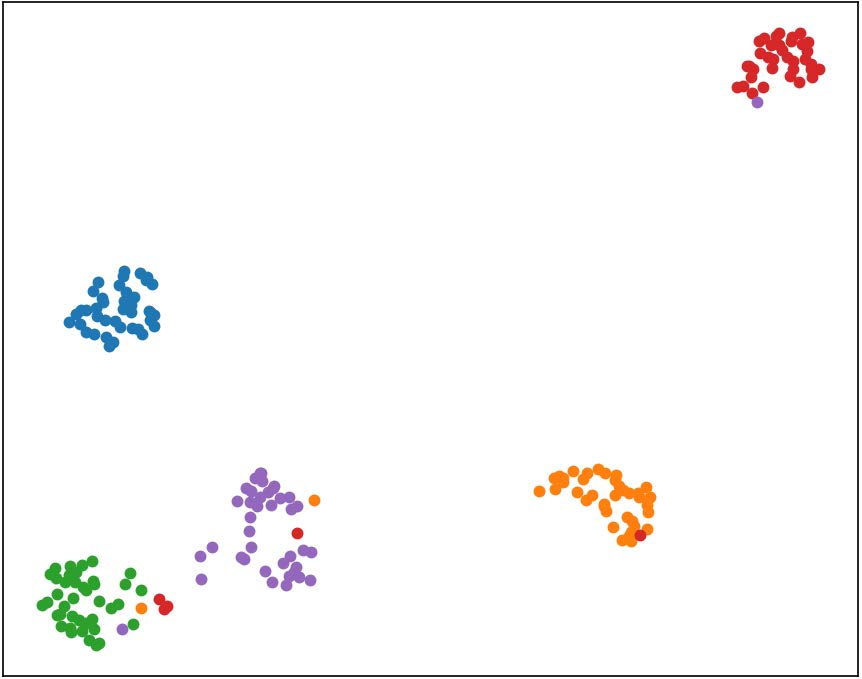}
    \subcaption{Kinetics}
    \label{fig6c}
  \end{subfigure}

  \vspace{0.8em}

  \hspace*{\fill}
  \begin{subfigure}[t]{0.3\linewidth}
    \centering
    \includegraphics[width=\linewidth]{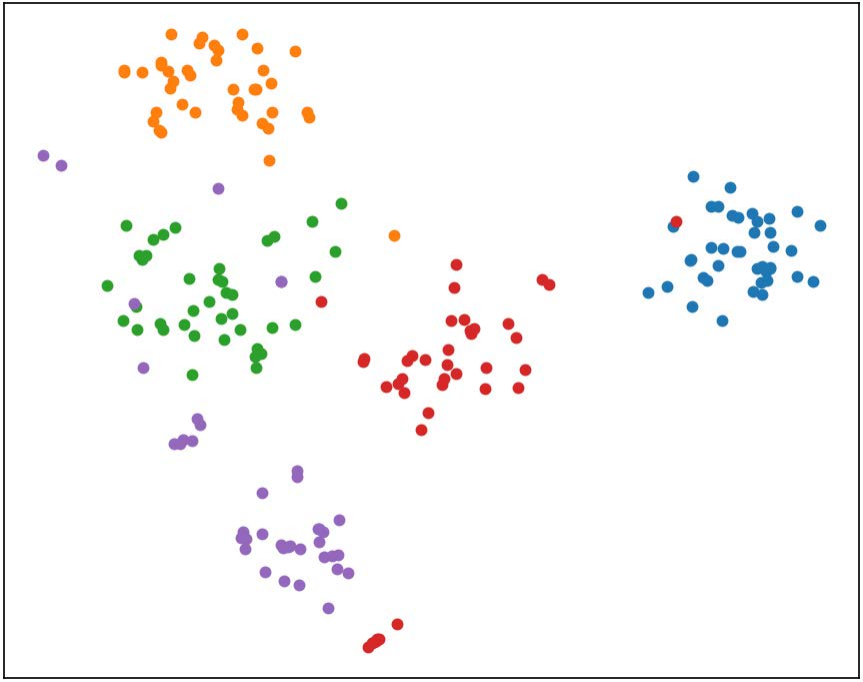}
    \subcaption{SSv2-FULL}
    \label{fig6d}
  \end{subfigure}
  \hfill
  \begin{subfigure}[t]{0.3\linewidth}
    \centering
    \includegraphics[width=\linewidth]{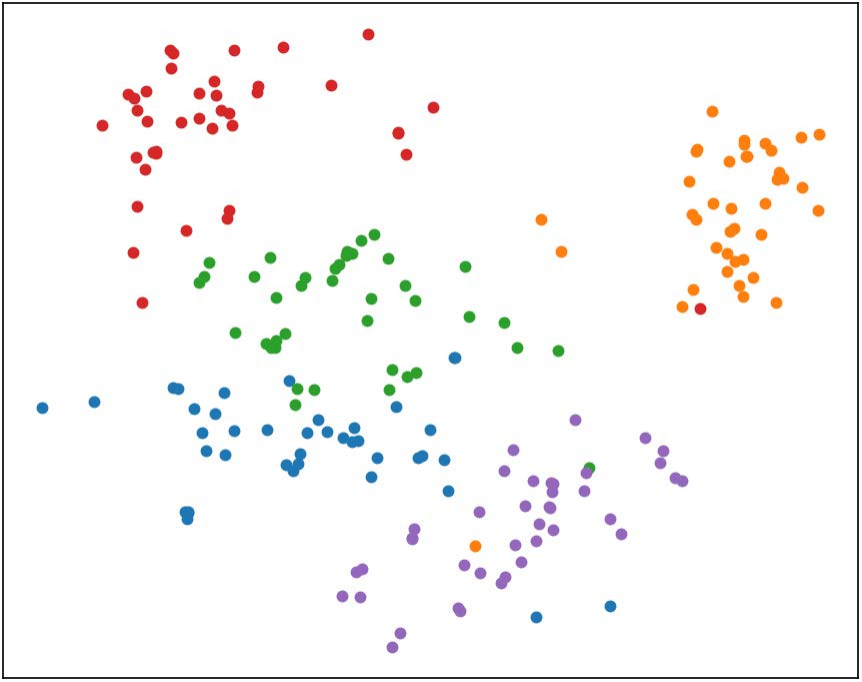}
    \subcaption{SSv2-Small}
    \label{fig6e}
  \end{subfigure}
  \hspace*{\fill}

  \caption{Distribution comparison on the test sets of HMDB51, UCF101, Kinetics, SSv2-FULL, and SSv2-Small. Five classes are shown, with 40 samples per class. Each dot denotes a sample. Colors indicate class labels.}
  \label{fig6}
\end{figure*}

\begin{figure*}[t]
  \centering

  \begin{subfigure}[t]{0.45\linewidth}
    \centering
    \includegraphics[width=\linewidth]{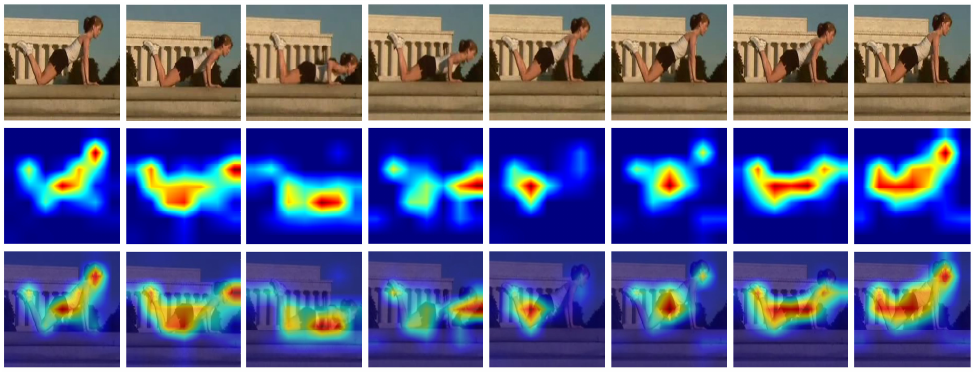}
    \subcaption{``Pushup'' in HMDB51}
    \label{fig7a}
  \end{subfigure}
  \hfill
  \begin{subfigure}[t]{0.45\linewidth}
    \centering
    \includegraphics[width=\linewidth]{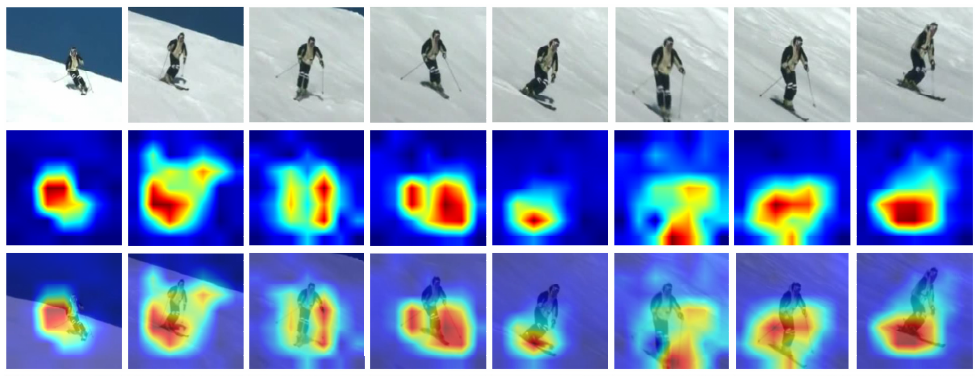}
    \subcaption{``Skiing'' in UCF101}
    \label{fig7b}
  \end{subfigure}

  \vspace{0.8em}

  \begin{subfigure}[t]{0.45\linewidth}
    \centering
    \includegraphics[width=\linewidth]{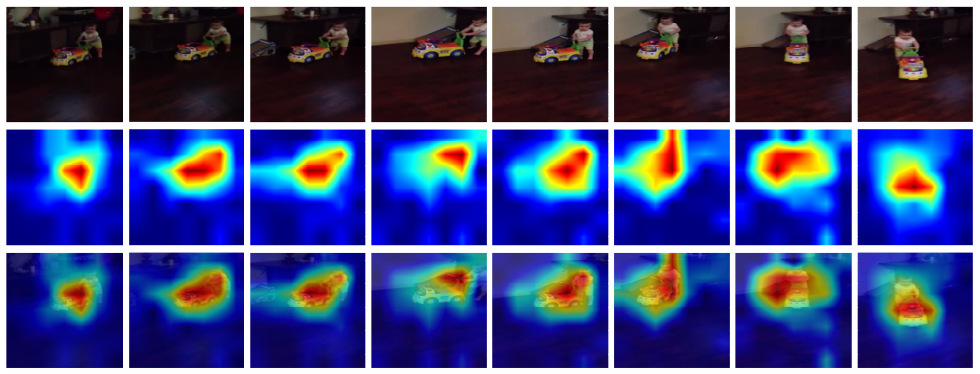}
    \subcaption{``Pushing car'' in Kinetics}
    \label{fig7c}
  \end{subfigure}
  \hfill
  \begin{subfigure}[t]{0.45\linewidth}
    \centering
    \includegraphics[width=\linewidth]{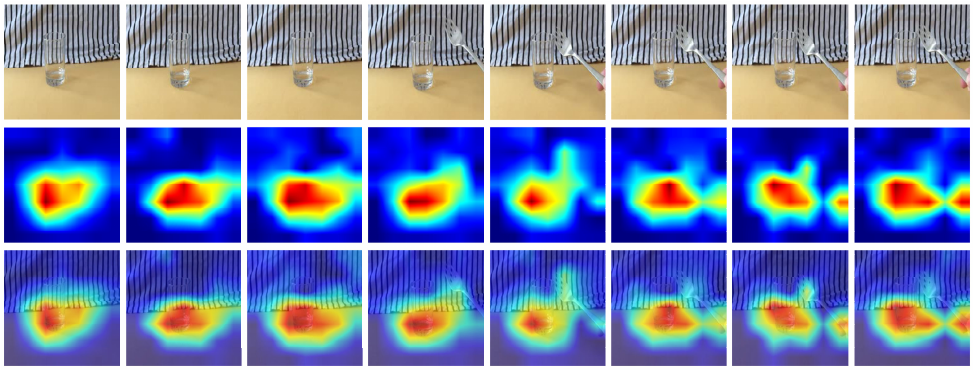}
    \subcaption{``Showing something next to something'' in SSv2-FULL}
    \label{fig7d}
  \end{subfigure}

  \vspace{0.8em}

  \begin{subfigure}[t]{0.45\linewidth}
    \centering
    \includegraphics[width=\linewidth]{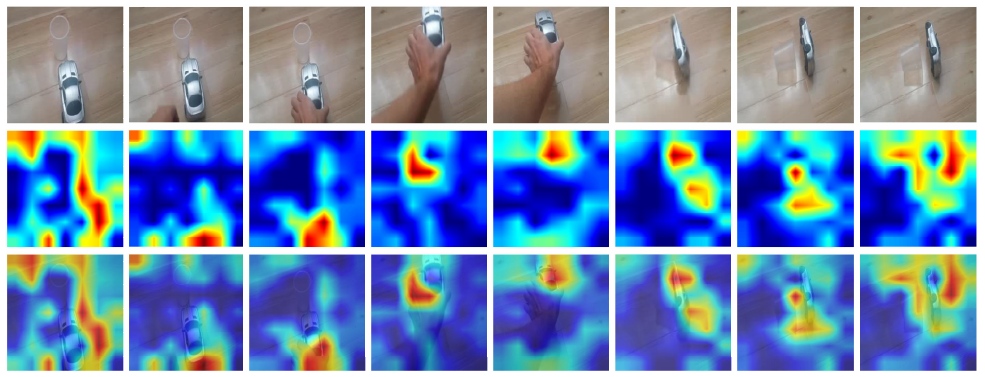}
    \subcaption{``Putting something on the edge of something so it is not supported and falls down'' in SSv2-Small}
    \label{fig7e}
  \end{subfigure}

  \caption{Visualization of the regions attended by CLIP-SPM during prediction. Each subfigure contains the original video frame, the attention heatmap, and their overlay, illustrating how the model focuses on action-relevant regions across different examples.}
  \label{fig7}
\end{figure*}

\subsubsection{Number of prompt templates}

We investigate the impact of the number of prompt templates by varying $R \in \{1,4,7,10,13,16\}$. As illustrated in Fig.~\ref{fig5}, performance exhibits a generally increasing trend on both HMDB51 and SSv2-Small as $R$ grows: accuracy on HMDB51 improves from 76.1\% to 77.4\%, and on SSv2-Small from 50.3\% to 50.8\%. Although slight local drops appear at some intermediate values---likely due to individual templates with weaker cross-episode generalization---the overall upward trajectory suggests that a richer prompt set provides more stable semantic cues for SPM and, in turn, yields more robust cross-modal modulation.

\subsection{Visualization and analysis}\label{subsec4.5}

To further understand the behavior of CLIP-SPM, we employ two visualization methods---UMAP~\cite{mcinnes2018umap} and Grad-CAM~\cite{selvaraju2017grad}---to examine the learned representations and the model's attention patterns.

We begin by analyzing the feature distributions using UMAP to assess inter-class separability and intra-class compactness. Fig.~\ref{fig6} shows the two-dimensional embeddings for HMDB51, UCF101, Kinetics, SSv2-FULL, and SSv2-Small. Each point corresponds to a test sample, with colors indicating class labels. The visualization reveals that CLIP-SPM produces highly discriminative representations on appearance-centric datasets such as UCF101, where classes form distinct and tightly clustered groups. HMDB51 and Kinetics display similarly clear inter-class separation, indicating strong discriminative capacity. In contrast, SSv2-FULL and SSv2-Small---both characterized by subtle motion variations and reduced reliance on appearance cues---exhibit more dispersed intra-class structures. Despite this increased difficulty, CLIP-SPM maintains well-defined inter-class boundaries, illustrating its robustness in learning motion representation even under higher temporal complexity.

We also examine the spatial focus of CLIP-SPM using Grad-CAM to visualize activation maps over input frames. Fig.~\ref{fig7} presents representative examples, highlighting the regions that most influence the model's predictions. The heatmaps consistently align with action-defining cues: limb and body articulation for ``Pushup'' and ``Skiing''; contact dynamics and motion direction for ``Pushing car''; object-object interactions for ``Showing something next to something''; and support boundary changes for ``Putting something on the edge of something so it is not supported and falls down.'' These observations demonstrate that CLIP-SPM attends to the correct spatiotemporal regions and captures fine-grained action-relevant semantics, confirming that its feature-modulation mechanisms guide attention toward the most informative portions of the visual input.

\subsection{Limitation and future work}\label{subsec4.6}

A current limitation of CLIP-SPM lies in its reliance on manually designed prompt templates, whose generalizability can vary across datasets. As illustrated in Fig.~\ref{fig5}, certain templates may introduce unstable semantic cues, occasionally resulting in small local drops in performance. Developing more principled or fully learnable prompt formulation strategies therefore represents a promising direction for future work, potentially improving the stability and robustness of semantic modulation across diverse tasks and domains.

\section{Conclusion}\label{sec5}

This paper introduced CLIP-SPM, a cross-modal framework for few-shot action recognition that addresses three fundamental challenges: robust temporal modeling, fine-grained visual similarity, and the modality gap between support and query samples. The proposed HSMR module enforces consistency between shallow and deep motion cues, enhancing temporal robustness; SPM learns query-relevant prompts and integrates them with visual features to enable effective vision-language interaction and improved semantic discriminability; and PADM jointly refines support prototypes and aligns query representations through a global anchor, strengthening cross-set consistency. Extensive experiments on five benchmark datasets demonstrate that CLIP-SPM consistently improves performance across 1-shot, 3-shot, and 5-shot settings, achieving competitive results compared with representative methods. Future work may explore more advanced prompt engineering methods to further enhance semantic modulation.

\appendix
\section{Prompt template}
\label{app1}
We adopt the 16 prompt templates listed in Table~\ref{tab6} to generate text features for each action class [CLS] in our experiments.
\begin{table}[htb]
  \centering
  \footnotesize
  \caption{The list of 16 different prompt templates.}
  \label{tab6}
  \begin{tabular}{l}
    \toprule
    Prompt template \\
    \midrule
    ``A photo of action [CLS]'' \\
    ``A picture of action [CLS]'' \\
    ``Human action of [CLS]'' \\
    ``[CLS], an action'' \\
    ``[CLS] this is an action'' \\
    ``[CLS], a video of action'' \\
    ``Playing action of [CLS]'' \\
    ``[CLS]'' \\
    ``Playing a kind of action, [CLS]'' \\
    ``Doing a kind of action, [CLS]'' \\
    ``Look, the human is [CLS]'' \\
    ``Can you recognize the action of [CLS]?'' \\
    ``Video classification of [CLS]'' \\
    ``A video of [CLS]'' \\
    ``The man is [CLS]'' \\
    ``The woman is [CLS]'' \\
    \bottomrule
  \end{tabular}
\end{table}

\bibliographystyle{main}
\bibliography{main}

\end{document}